\pdfoutput=1

\documentclass[11pt]{article}

\usepackage[]{ACL2023}

\usepackage{times}
\usepackage{latexsym}

\usepackage[T1]{fontenc}

\usepackage[utf8]{inputenc}

\usepackage{microtype}

\usepackage{inconsolata}

\usepackage{MnSymbol, textcomp}
\usepackage{amsfonts, amsmath}
\usepackage{adjustbox, graphics}
\usepackage{array, booktabs, caption, multirow, pbox, subcaption}
\usepackage{tikz-qtree, tikz-qtree-compat}
\usepackage{soul, xcolor}
\usepackage{gb4e}
\noautomath

\DeclareUnicodeCharacter{2070}{\ensuremath{{}^0}}

\newcommand{\PreserveBackslash}[1]{\let\temp=\\#1\let\\=\temp}
\newcolumntype{C}[1]{>{\PreserveBackslash\centering}p{#1}}
\newcolumntype{R}[1]{>{\PreserveBackslash\raggedleft}p{#1}}
\newcolumntype{L}[1]{>{\PreserveBackslash\raggedright}p{#1}}

\tikzset{every tree node/.style={align=center, anchor=north}}

\definecolor{forestgreen}{RGB}{34, 139, 34}
\definecolor{darkred}{RGB}{139, 0, 0}
\definecolor{lightpink}{RGB}{254, 189, 208}
\definecolor{surf}{RGB}{197, 222, 203}
\sethlcolor{lightpink}

\title{
    Bring More Attention to Syntactic Symmetry \\
    for Automatic Postediting of High-Quality Machine Translations
}

\author{
  Baikjin Jung$^\diamondsuit$ \quad Myungji Lee$^\heartsuit$ \quad Jong-Hyeok Lee$^{\diamondsuit \heartsuit}$ \quad Yunsu Kim$^{\diamondsuit \heartsuit}$ \\ \\
  $^\diamondsuit$Department of Computer Science and Engineering \\
  $^\heartsuit$Graduate School of Artificial Intelligence \\
  Pohang University of Science and Technology, Republic of Korea \\
  \texttt{\{bjjung, mjlee7, jhlee, yunsu.kim\}@postech.ac.kr}
}

\begin{document}
\maketitle

\begin{abstract}

Automatic postediting (APE) is an automated process to refine a given machine translation (MT).
Recent findings present that existing APE systems are not good at handling high-quality MTs even for a language pair with abundant data resources, English--German: the better the given MT is, the harder it is to decide what parts to edit and how to fix these errors.
One possible solution to this problem is to instill deeper knowledge about the target language into the model.
Thus, we propose a linguistically motivated method of regularization that is expected to enhance APE models' understanding of the target language: a loss function that encourages symmetric self-attention on the given MT.
Our analysis of experimental results demonstrates that the proposed method helps improving the state-of-the-art architecture's APE quality for high-quality MTs.

\end{abstract}

\section{Introduction}

Automatic postediting (APE) is an automated process to transform a given machine translation (MT) into a higher-quality text~\cite{knight-chander-1994}.
Since 2015, Conference on Machine Translation (WMT) has been hosting an annual shared task for APE, and most of the recently developed APE systems are within the common framework of representation learning using artificial neural networks to learn postediting patterns from the training data~\cite{chatterjee-etal-2018, chatterjee-etal-2019, chatterjee-etal-2020, akhbardeh-etal-2021}.

Since 2018, all participants in the shared task have used Transformer-based models~\cite{vaswani-etal-2017}, but recent findings of the shared task~\cite{chatterjee-etal-2018, chatterjee-etal-2019, chatterjee-etal-2020, akhbardeh-etal-2021} cast doubt on whether Transformer-based APE models learn good generalizations because such models' APE quality appears to be significantly affected by external factors such as the source--target language pair, the qualitative characteristics of the provided data, and the quality of the given MT.

Especially, the good quality of the given MTs has brought great difficulty in performing APE on the WMT 2019 test data set: the better the given MT is, the harder it is to decide what parts to edit and how to correct these errors~\cite{chatterjee-etal-2018, chatterjee-etal-2019}.
The thing to notice is that this outcome is not a question of data scarcity because the language pair of this test data set, English--German, is a language pair provided with abundant training, validation, and test data.
Also, it is not a question of data heterogeneity, either: the domain of this test data set, IT, shows a high degree of lexical repetition, which indicates that data sets in this domain use the same small set of lexical items~\cite{chatterjee-etal-2018, chatterjee-etal-2019, akhbardeh-etal-2021}.
Thus, it would be a question of modeling, and one possible solution is to implant deeper knowledge about the target language into the model.

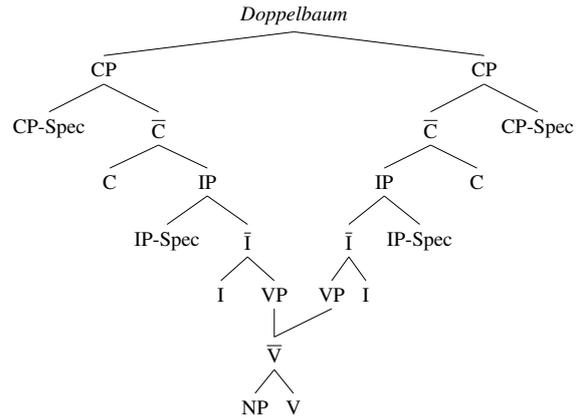
\begin{figure}[t]
\centering
\adjustbox{width=\columnwidth}{
    \begin{tikzpicture}
    \Tree
    [.\textit{Doppelbaum}
        [.CP
            [.CP-Spec ]
            [.{$\overline{\mbox{C}}$}
                [.C ]
                [.IP
                    [.IP-Spec ]
                    [.{$\overline{\mbox{I}}$}
                        [.I ]
                        [.VP
                            [.\node (1) {$\overline{\mbox{V}}$};
                                [.NP ]
                                [.V ]
                            ]
                        ]
                    ]
                ]
            ]
        ]
        [.CP
            [.{$\overline{\mbox{C}}$}
                [.IP
                    [.{$\overline{\mbox{I}}$}
                        [.\node (2) {VP}; ]
                        [.I ]
                    ]
                    [.{IP-Spec} ]
                ]
                [.C ]
            ]
            [.CP-Spec ]
        ]
    ]
    \draw (2.south) -- (1.north);
    \end{tikzpicture}
}
\caption{
    A depiction of \textit{Doppelbaum} (\S\ref{sec:feld}).
}
\label{fig:doppelbaum}
\end{figure}

To this end, we propose a new method of regularization that is expected to enhance Transformer-based APE models' understanding of German translations.
Specifically, the proposed method is based on \textit{Feldermodell} (\S \ref{sec:feld}), an established linguistic model, which implies the need for proper treatment of the underlying symmetry of German sentence structures.
To instill the idea of syntactic symmetry into Transformer-based APE models, we introduce a loss function that encourages symmetric self-attention on the given MT.
Based on experimental results, we conduct a careful analysis and conclude that the proposed method has a positive effect on improving the state-of-the-art architecture's APE quality for high-quality MTs.

\section{Linguistic Theory}
\label{sec:feld}

In German linguistics, \textit{das topologische Satzmodell} (`the topological sentence model') or \textit{das Feldermodell} (`the field model')~\cite{reis-1980, woellstein-2018, hoehle-2019} describes how constituents of a sentence are closely related even if they are far apart from each other.
Usually, \textit{Feldermodell} divides a clause into \textit{das Vorfeld} (`the prefield'; VF), \textit{die linke Satzklammer} (`the left bracket'; LSK), \textit{das Mittelfeld} (`the middlefield'; MF), \textit{die rechte Satzklammer} (`the right bracket'; RSK), and \textit{das Nachfeld} (`the postfield'; NF).

\begin{exe}
    \ex $\bigl[$ Heute $_{\text{VF}}\bigr]$ $\bigl[$ habe $_{\text{LSK}}\bigr]$ $\bigl[$ ich $_{\text{MF}}\bigr]$ $\bigl[$ gesehen $_{\text{RSK}}\bigr]$ $\bigl[$ zuf{\"a}llig $_{\text{NF}}\bigr]$, \label{sent:1}
    \ex $\Bigl[$ $\bigl[$ dass $_{\text{LSK}}\bigr]$ $\bigl[$ du eine Tasse Kaffee $_{\text{MF}}\bigr]$ $\bigl[$ getrunken hast $_{\text{RSK}}\bigr]$ $_{\text{NF}}\Bigr]$. \label{sent:2}
\end{exe}

These parts are all interrelated; LSK and RSK are a typical example: while the former holds a finite verb or a complementizer, the latter holds a past participle, an infinitive, and a particle.
In (\ref{sent:1}), VF holds ``\textit{Heute}'' (`today'); LSK holds ``\textit{habe}'' (`have'); MF holds ``\textit{ich}'' (`I'); RSK holds ``\textit{gesehen}'' (`seen'); and NF holds ``\textit{zuf{\"a}llig}'' (`by chance').
(\ref{sent:2}) is an additional NF of (\ref{sent:1}) and includes its own LSK holding ``\textit{dass}'' (`that'); MF holding ``\textit{du eine Tasse Kaffee}'' (`you a cup of coffee'); and RSK holding ``\textit{getrunken hast}'' (`drank').

For such analyses, special tree structures such as \textit{Doppelbaum}~\cite{woellstein-2018} (`double tree') can be used, which is a bimodal tree (Fig.~\ref{fig:doppelbaum}), where two CP, $\overline{\mbox{C}}$, IP, $\overline{\mbox{I}}$, and VP subtrees are `\textbf{symmetric}' with respect to $\overline{\mbox{V}}$.
We assume that this structural symmetry is parameterized from the perspective, not only of generative linguistics~\cite{woellstein-2018, hoehle-2019}, but also of a parametric model $\mathcal{P} = \{ P_\theta \mid \theta \in \Theta \}$, where $P_\theta$ and $\Theta$ are a probability distribution and the parameter space, respectively.

Especially, if we look at APE in terms of sequence-to-sequence learning~\cite{sutskever-etal-2014}, the probability distribution of the output sequence $(y_{1}, \cdots, y_{L_{y}})$ is obtained in the following manner:
\begin{align*}
    & P_{\theta}(y_{1}, \cdots, y_{L_{y}} \mid x_{1}, \cdots, x_{L_{x}}, z_{1}, \cdots, z_{L_{z}}) \\
    & = \prod_{t=1}^{L_{y}} P_{\theta}(y_{t} \mid u, v, y_{1}, \cdots, y_{t-1})\text{,}
\end{align*}
where $u$ and $v$ are the representations of a source text $(x_{1}, \cdots, x_{L_{x}})$ and its MT $(z_{1}, \cdots, z_{L_{z}})$, respectively.
In this process, we presume that the syntactic symmetry of the target language affects the resulting distribution $P_{\theta}$; in other words, this syntactic symmetry would be an inductive bias~\cite{mitchell-1980} that should be handled properly.

\section{Methodology}
\label{sec:method}

We implement a multi-encoder Transformer model consisting of the ``Joint-Final'' encoder and the ``Parallel'' decoder, which is a state-of-the-art architecture for APE~\cite{shin-etal-2021}, and conduct a controlled experiment without concern for usage of performance-centered tuning techniques.
Specifically, the Joint-Final encoder consists of a source-text encoder and an MT encoder, which process the given source text and MT, respectively.
Based on this baseline architecture, we propose a method to encourage the MT encoder to perform symmetric self-attention by minimizing the skewness of each self-attention layer's categorical distribution $p_{\text{self}}$.

The used measure of skewness is
\begin{equation*}
(\ddot{\mu}_{3})_{i} = \left( \sum_{j=1}^{\lfloor \frac{L_{z}}{2} \rfloor} p_{\text{self}}[i, j] - \sum_{j=\lceil \frac{L_{z}}{2} \rceil + 1}^{L_{z}} p_{\text{self}}[i, j] \right)^{2}\text{,}
\label{eq:skewness}
\end{equation*}
for each token $z_{i}$ in the given MT $(z_{1}, \cdots, z_{L_{z}})$.

Accordingly, the basic cross-entropy loss $\mathcal{L}_{\text{CE}}$ is regularized by $(\ddot{\mu}_{3})_{i}$, resulting in a new loss function
\begin{equation*}
\mathcal{L}_{\textsc{Doppelbaum}} = \mathcal{L}_{\text{CE}} + \mathbb{E} \big[ \alpha \big] \mathbb{E} \big[ \ddot{\mu}_{3} \big] + (1 - \alpha)\text{,}
\end{equation*}
where
\begin{equation*}
\mathbb{E} \big[ \alpha \big] = \frac{\sum_{b=1}^{B} \sum_{i=1}^{L_{z}} \alpha_{b, i}}{B \times L_{z}}
\end{equation*}
is the expected value of coefficients
\begin{equation*}
\alpha_{b, i} = \sigma(W^{\text{T}} v_{b, i} + \beta)
\end{equation*}
in the given minibatch, and
\begin{equation*}
\mathbb{E} \big[ \ddot{\mu}_{3} \big] = \frac{\sum_{b=1}^{B} \sum_{n=1}^{N} \sum_{h=1}^{H} \sum_{i=1}^{L_{z}} (\ddot{\mu}_{3})_{b, n, h, i}}{B \times N \times H \times L_{z}}
\end{equation*}
is the expected value of $(\ddot{\mu}_{3})_{b, n, h, i}$.
In addition, $(1 - \alpha)$ is an initial inducement to utilizing $\ddot{\mu}_{3}$.
In the equations above, $\sigma$ is the sigmoid function, $v$ is the output of the final layer of the MT encoder, $W \in \mathbb{R}^{d_{\text{model}}}$ and $\beta \in \mathbb{R}$ are learned parameters, $B$ is the number of data examples, $N$ is the number of layers, and $H$ is the number of heads.

\section{Experiment}

In the conducted experiment, all hyperparameters are the same as those of \citet{shin-etal-2021} except the learning rate (Appendix~\ref{sec:exp}); we basically reproduce their experimental design.

\begin{table}[ht]
\centering
\small

\begin{tabular}{
    C{0.25\columnwidth} C{0.43\columnwidth} R{0.16\columnwidth}
}
    \toprule
    \multicolumn{2}{c}{\textsc{Data Sets}} & \multicolumn{1}{c}  {\textsc{Sizes}} \\
    \midrule
    \multirow{2}{*}{\textsc{Training}} & eSCAPE-NMT & 5,065,187 \\
    & WMT 2019 & 13,442 \\
    \midrule
    \textsc{Validation} & WMT 2019 & 1,000 \\
    \midrule
    \textsc{Test} & WMT 2019 & 1,023 \\
    \bottomrule
\end{tabular}

\caption{
    APE data sets used in the experiment.
    eSCAPE-NMT is a cleansed subset of \href{http://hltshare.fbk.eu/QT21/eSCAPE.html}{eSCAPE}'s~\cite{negri-etal-2018} English--German-NMT set.
    The cleansing procedure is a reproduction of \citet{shin-etal-2021}.
    The WMT 2019 data sets~\cite{chatterjee-etal-2019} were released for WMT 2018 but used also at WMT 2019.
    }
\label{tab:data}
\end{table}

Both the baseline model and the proposed model are trained by using the training data sets and the validation data set listed in Table~\ref{tab:data}; we first train the models by using eSCAPE-NMT mixed with the WMT 2019 training data in the ratio of $27:1$, and then tune them by using the WMT 2019 training data solely.

\section{Results and Analysis}
\label{sec:analysis}

\begin{table}[ht]
\centering
\small

\begin{tabular}{
    C{0.32\columnwidth} L{0.08\columnwidth} L{0.15\columnwidth} L{0.08\columnwidth} L{0.1\columnwidth}
}
    \toprule

    \multirow{2}{*}{
        \pbox[c]{1\linewidth}{\vspace{0.5em} \textsc{Systems}}
    } & \multicolumn{4}{c}{\textsc{WMT 2019}} \\

    \cmidrule(lr){2-5}

    & \multicolumn{2}{c}{TER\textsuperscript{\textdownarrow} ($\sigma$)} & \multicolumn{2}{c}{\textsc{Bleu}\textsuperscript{\textuparrow} ($\sigma$)} \\

    \midrule

    Given MT & 16.84 & (19.52) & 74.73 & (25.89) \\
    Baseline & 16.60\textsuperscript{\dagger} & (19.51) & 75.11\textsuperscript{\dagger} & (26.21) \\

    \textsc{Doppelbaum} & \textbf{16.54}\textsuperscript{\dagger} & (19.48) & \textbf{75.47}\textsuperscript{\dagger}\textsuperscript{*} & (26.16) \\

    \bottomrule
\end{tabular}

\caption{
    The results of automatic evaluation on the WMT 2019 test data set.
    Baseline is the above-mentioned baseline model (\S \ref{sec:method}), and \textsc{Doppelbaum} is the proposed model.
    Beside TER~\cite{snover-etal-2006} and \textsc{Bleu}~\cite{papineni-etal-2002}, their sentence-level standard deviations ($\sigma$) are presented.
    In each column, the figure implying the best performance is in \textbf{bold}.
    The dagger symbols denote the proposed model's quality improvement on the given MTs is statistically significant ($p \leq 0.05$).
    The asterisks denote the proposed model's performance improvement on the baseline model is statistically significant ($p \leq 0.05$).
}
\label{tab:result}
\end{table}

\begin{figure}[ht]
\centering
\includegraphics[width=\columnwidth]{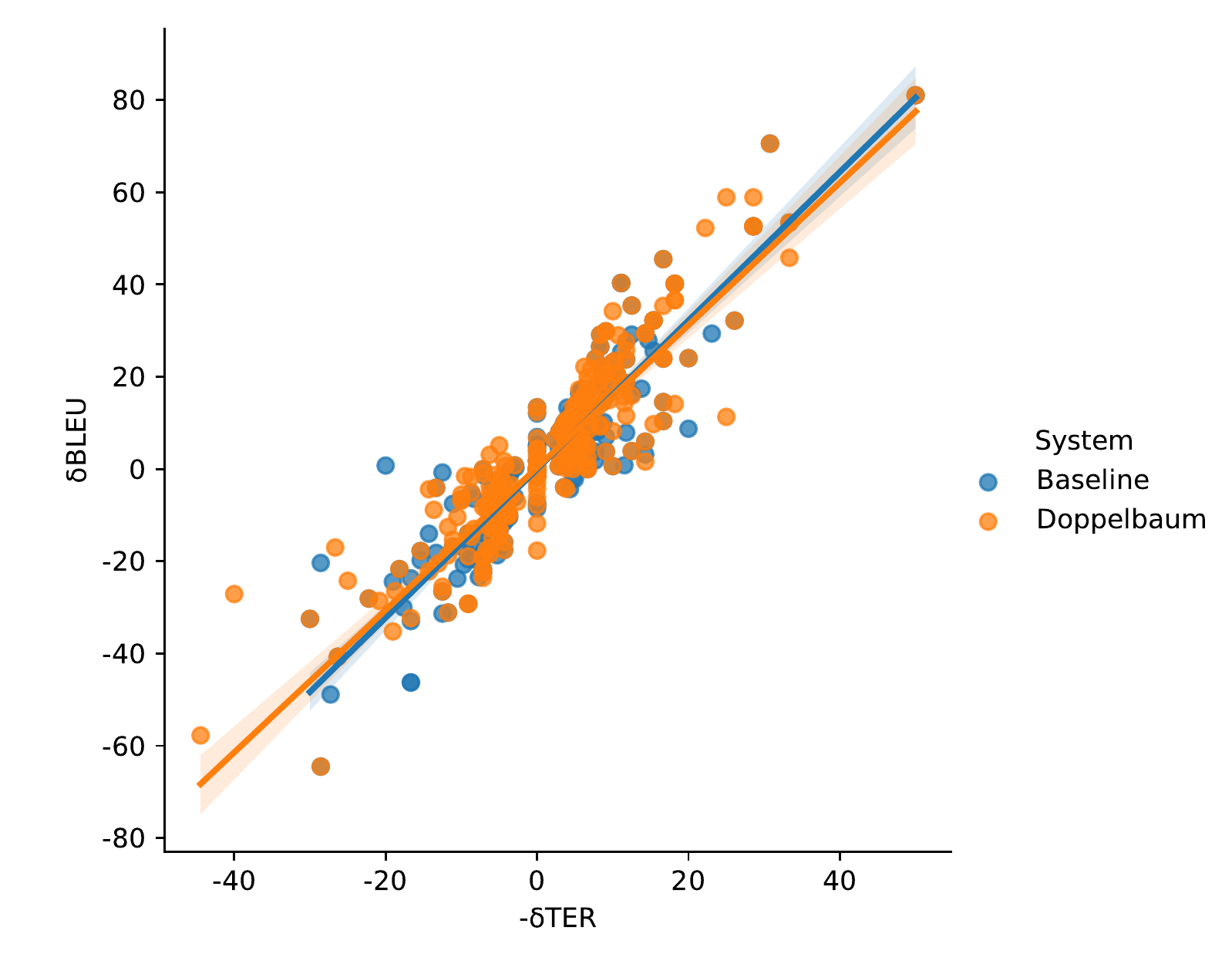}
\caption{
    The relationship between models' sentence-level TER improvements ($-\delta$TER; positive values denote decrease in TER) and sentence-level \textsc{Bleu} improvements ($\delta$\textsc{Bleu}; positive values denote increase in \textsc{Bleu}) on those of the given MTs in the test data set.
}
\label{fig:linear}
\end{figure}

\begin{table*}[t]
\centering
\small

\begin{tabular}{
    C{0.16\textwidth} C{0.06\textwidth} R{0.06\textwidth} R{0.06\textwidth} R{0.06\textwidth} R{0.06\textwidth} R{0.06\textwidth} R{0.06\textwidth} R{0.06\textwidth} C{0.095\textwidth}
}
    \toprule

    \multirow{2}{*}{
        \pbox[c]{1\linewidth}{\vspace{0.5em} \textsc{Systems}}
    } & & \multicolumn{5}{c}{\textsc{Modified}} & \multicolumn{2}{c}{\textsc{Intact}} & \multirow{2}{*}{
        \pbox[c]{1\linewidth}{\vspace{0.5em} \textsc{F1}}
    } \\
    
    \cmidrule(lr){3-7} \cmidrule(lr){8-9}
    
    & & \textcolor{darkred}{\textsc{Ruin}} & \textcolor{darkred}{\textsc{Degr}} & \textsc{Even} & \textcolor{forestgreen}{\textsc{Impr}} & \textcolor{forestgreen}{\textsc{Perf}} & \textcolor{forestgreen}{\textsc{Acce}} & \textcolor{darkred}{\textsc{Negl}} & \\
    
    \midrule
    \\[-2ex]
    
    \multirow{3}{*}{Baseline} & \% & 1.86 & \textbf{6.65} & 5.67 & \textbf{7.14} & 4.30 & 23.36 & 51.03 & \multirow{3}{*}{22.8} \\
    & $\mu_{\delta \textsc{Bleu}}$ & $-$24.48 & $-$13.51 & 0.50 & 9.22 & 27.23 & 0.00 & 0.00 & \\
    & $\sigma_{\delta \textsc{Bleu}}$ & 15.48 & 9.42 & 3.38 & 8.43 & 16.39 & 0.00 & 0.00 & \\

    \\[-1.5ex]
    \hline
    \\[-1.5ex]

    \multirow{3}{*}{\textsc{Doppelbaum}} & \% & \textbf{1.56} & 7.33 & 5.77 & \textbf{7.14} & \textbf{5.87} & \textbf{23.66} & \textbf{48.68} & \multirow{3}{*}{\textbf{25.4}} \\
    & $\mu_{\delta \textsc{Bleu}}$ & $-$26.12 & $-$11.72 & $-$0.42 & 10.04 & 27.21 & 0.00 & 0.00 & \\
    & $\sigma_{\delta \textsc{Bleu}}$ & 16.09 & 9.16 & 3.82 & 8.69 & 16.37 & 0.00 & 0.00 & \\

    \\[-2ex]
    \bottomrule

\end{tabular}

\caption{
    A relative frequency distribution containing the frequencies of the following groups (we compare the TER of the given MT and that of the postedited result.): the cases where an APE system injects errors to an already perfect MT (\textcolor{darkred}{\textsc{Ruin}}); both the given MT and the APE result are not perfect, but the former is better in terms of TER (\textcolor{darkred}{\textsc{Degr}}); both are not perfect and have the same TER although they are different from each other (\textsc{Even}); both are not perfect, but the latter is better (\textcolor{forestgreen}{\textsc{Impr}}); the given MT is not perfect whereas the APE result is (\textcolor{forestgreen}{\textsc{Perf}}); both are perfect (\textcolor{forestgreen}{\textsc{Acce}}); and lastly, even though the MT is not perfect, the APE system does not change anything (\textcolor{darkred}{\textsc{Negl}}).
    The calculation of the F1 score is based on two criteria: whether the given MT is perfect or not (for recall) and whether the APE system edits the given MT or not (for precision).
    \% is the proportion of the cases belonging to each category, $\mu_{\delta \textsc{Bleu}}$ is the average of sentence-level \textsc{Bleu} improvements, and $\sigma_{\delta \textsc{Bleu}}$ is their standard deviation.
}
\label{tab:frequencies}
\end{table*}

\begin{table}[!ht]
\centering
\small

\begin{tabular}{
    C{0.14\columnwidth} C{0.2\columnwidth} L{0.26\columnwidth} C{0.07\columnwidth} R{0.05\columnwidth}
}
    \toprule
    \multicolumn{3}{c}{\textsc{Types of APE}} & \multicolumn{2}{c}{\textsc{Numbers}} \\

    \midrule
    \\[-2ex]
    
    \multirow{8}{*}[-1.5ex]{\textsc{Perf}} & \multirow{4}{*}{Linguistic} & Nouns & \multicolumn{2}{r}{5} \\
    & & Expressions & \multicolumn{2}{r}{5} \\
    & & Agreement & \multicolumn{2}{r}{3} \\
    & & Prepositions & \multicolumn{2}{r}{2} \\

    \\[-1.5ex]
    \cmidrule(lr){2-5}
    \\[-1.5ex]

    & \multirow{3}{*}{Other} & Punctuation & \multicolumn{2}{r}{5} \\
    & & URLs & \multicolumn{2}{r}{2} \\
    & & Noise Removal & \multicolumn{2}{r}{2} \\

    \\[-2ex]
    \cmidrule(lr){2-5}
    \\[-2ex]

    & & & Total & 24 \\

    \\[-2ex]
    \midrule
    \\[-2ex]

    \multirow{5}{*}[-2.75ex]{\textsc{Acce}} & \multirow{3}{*}{Linguistic} & Nouns & \multicolumn{2}{r}{3} \\
    & & Expressions & \multicolumn{2}{r}{2} \\
    & & Adjectives & \multicolumn{2}{r}{1} \\

    \\[-1.5ex]
    \cmidrule(lr){2-5}
    \\[-1.5ex]

    & Other & Punctuation & \multicolumn{2}{r}{2} \\

    \\[-2ex]
    \cmidrule(lr){2-5}
    \\[-2ex]

    & & & Total & 8 \\

    \\[-2ex]
    \bottomrule

\end{tabular}

\caption{
    Manual categorization of the cases where only the proposed model produces a perfect translation.
    For more information on the definitions of \textsc{Perf} and \textsc{Acce}, refer to Table~\ref{tab:frequencies}.
    `Linguistic' and `Other' cases are results of nontrivial postediting and trivial postediting, respectively.
    `Expressions' means using appropriate determiners, verb phrases, shortened forms of the definite article, etc..
    `Noise Removal' means filtering out meaningless tokens from the given MT.
    This categorization was double-checked by a native German speaker.
}
\label{tab:ape-type}
\end{table}

\begin{table*}[ht]
\centering
\small

\begin{tabular}{
    C{0.18\textwidth} L{0.6\textwidth} R{0.05\textwidth} R{0.06\textwidth}
}
\toprule

\multicolumn{2}{c}{\textsc{Case 1: Perf}} & \multicolumn{1}{c}{TER\textsuperscript{\textdownarrow}} & \multicolumn{1}{c}{\textsc{Bleu\textsuperscript{\textuparrow}}} \\

\midrule
\\[-1ex]

Source Text & \pbox[c]{0.665\textwidth}{
    For example , the following function contains variables that are defined in \\
    various block scopes .
} & & \\
\\[-1ex]

Given MT & \pbox{0.665\textwidth}{
    \colorbox{surf}{Die} \colorbox{surf}{folgende} \colorbox{surf}{Funktion} \colorbox{surf}{enth{\"a}lt} \colorbox{surf}{zum} \colorbox{surf}{Beispiel} \colorbox{lightpink}{Variable} \colorbox{surf}{,} \colorbox{surf}{die} \\
    \colorbox{surf}{in} \colorbox{surf}{verschiedenen} \colorbox{surf}{Codebereichen} \colorbox{surf}{definiert} \colorbox{surf}{sind} \colorbox{surf}{.}
} & 6.67 & 80.03 \\

\\[-1ex]
\hline
\\[-1ex]

Baseline & \pbox[c]{0.665\textwidth}{
    \colorbox{surf}{Die} \colorbox{surf}{folgende} \colorbox{surf}{Funktion} \colorbox{surf}{enth{\"a}lt} \colorbox{surf}{zum} \colorbox{surf}{Beispiel} \colorbox{lightpink}{Variable} \colorbox{surf}{,} \colorbox{surf}{die} \\
    \colorbox{surf}{in} \colorbox{surf}{verschiedenen} \colorbox{surf}{Codebereichen} \colorbox{surf}{definiert} \colorbox{surf}{sind} \colorbox{surf}{.}
} & 6.67 & 80.03 \\
\\[-1ex]

\textsc{Doppelbaum} & \pbox[c]{0.665\textwidth}{
    \colorbox{surf}{Die} \colorbox{surf}{folgende} \colorbox{surf}{Funktion} \colorbox{surf}{enth{\"a}lt} \colorbox{surf}{zum} \colorbox{surf}{Beispiel} \colorbox{surf}{Variablen} \colorbox{surf}{,} \colorbox{surf}{die} \\
    \colorbox{surf}{in} \colorbox{surf}{verschiedenen} \colorbox{surf}{Codebereichen} \colorbox{surf}{definiert} \colorbox{surf}{sind} \colorbox{surf}{.}
} & 0.00 & 100.00 \\

\\[-1ex]
\hline
\\[-1ex]

Manual Postediting & \pbox[c]{0.665\textwidth}{
    Die folgende Funktion enth{\"a}lt zum Beispiel Variablen , die in \\
    verschiedenen Codebereichen definiert sind .
} & & \\

\\[-1ex]
\bottomrule

\end{tabular}

\caption{
    A case where only the proposed model corrects the given MT perfectly.
    Considering the manually postedited result, wrong words in the given MT, the APE result of the baseline model, and that of the proposed model are highlighted in pink while correct words are highlighted in green.
    All the texts are tokenized or detokenized using Moses~\cite{koehn-etal-2007}.
}
\label{tab:case-1}
\end{table*}

\begin{table*}[ht]
\centering
\small

\begin{tabular}{
    C{0.18\textwidth} L{0.6\textwidth} R{0.05\textwidth} R{0.06\textwidth}
}
\toprule

\multicolumn{2}{c}{\textsc{Case 2: Acce}} & \multicolumn{1}{c}{TER\textsuperscript{\textdownarrow}} & \multicolumn{1}{c}{\textsc{Bleu\textsuperscript{\textuparrow}}} \\

\midrule
\\[-1ex]

Source Text & \pbox[c]{0.665\textwidth}{
    Double-click the Zoom tool .
} & & \\
\\[-1ex]

Given MT & \pbox{0.665\textwidth}{
    \colorbox{surf}{Doppelklicken} \colorbox{surf}{Sie} \colorbox{surf}{auf} \colorbox{surf}{das} \colorbox{surf}{Zoomwerkzeug} \colorbox{surf}{.}
} & 0.00 & 100.00 \\

\\[-1ex]
\hline
\\[-1ex]

Baseline & \pbox[c]{0.665\textwidth}{
    \colorbox{surf}{Doppelklicken} \colorbox{surf}{Sie} \colorbox{surf}{auf} \colorbox{surf}{das} \colorbox{lightpink}{Zoom-Werkzeug} \colorbox{surf}{.}
} & 16.67 & 53.73 \\
\\[-1ex]

\textsc{Doppelbaum} & \pbox[c]{0.665\textwidth}{
    \colorbox{surf}{Doppelklicken} \colorbox{surf}{Sie} \colorbox{surf}{auf} \colorbox{surf}{das} \colorbox{surf}{Zoomwerkzeug} \colorbox{surf}{.}
} & 0.00 & 100.00 \\

\\[-1ex]
\hline
\\[-1ex]

Manual Postediting & \pbox[c]{0.665\textwidth}{
    Doppelklicken Sie auf das Zoomwerkzeug .
} & & \\

\\[-1ex]
\bottomrule

\end{tabular}

\caption{
    A case where only the proposed model adopts the given, already perfect MT. Details are the same as in Table~\ref{tab:case-1}.
}
\label{tab:case-2}
\end{table*}

The result of automatic evaluation (Table~\ref{tab:result}) indicates that the proposed model improves on the baseline model in terms of \textsc{Bleu} (75.47) but does not in terms of TER (16.54), which is unusual.
Although those measures have a strong correlation overall (Fig.~\ref{fig:linear}), the proposed model has more outliers, $\delta$\textsc{Bleu} (the value obtained by subtracting a given MT's \textsc{Bleu} from the postedited result's \textsc{Bleu}) of which is over 20, compared to the baseline model; they must be the ones that bring the improvement in \textsc{Bleu}.

Thus, we present an additional evaluation result to further investigate this mismatch between TER improvements and \textsc{Bleu} improvements: a relative frequency distribution of successes and failures in APE with regard to the TER difference between a given MT and each model's output (Table~\ref{tab:frequencies}).
Then, the mentioned outliers correspond to \textsc{Perf}, which is the set of the cases where an APE system succeeds in perfectly correcting the given MT with one or more errors, considering that the proposed model's \textsc{Perf} has a $\mu_{\delta \textsc{Bleu}}$ (the average of sentence-level \textsc{Bleu} improvements) of 27.21.
We see that the proposed model has substantially more \textsc{Perf} cases (5.87\%) than the baseline model (4.30\%) and that because most of those `new' (1.57\textit{pp}) cases are results of nontrivial postediting (Table~\ref{tab:ape-type}), this increase in the proportion of perfect postediting is valid evidence of the proposed method's effect on enhancing the baseline model's APE quality for high-quality MTs.

In addition, in an actual example where only the proposed model corrects the given MT perfectly (Table~\ref{tab:case-1}), we observe that the proposed model successfully captures the close relation between the verb ``\textit{enth{\"a}lt}'' (`contains') and its object so that the correct form ``\textit{Variablen}'' (`variables') is used.
Considering that the adverb phrase ``\textit{zum Beispiel}'' (`for example') in the given MT makes some distance between the verb and its object, it appears that the proposed model integrates information from a wider range of constituents than the baseline model; hence the conclusion that the proposed method instills \textit{Feldermodell}'s idea of syntactic symmetry into Transformer-based APE models and enhances their understanding of German translations.

Another example (Table~\ref{tab:case-2}) suggests that the increase in the proportion of \textsc{Acce} (0.3\textit{pp}), which is the set of the cases where an APE system adopts the given, already perfect MT, should be cautiously interpreted.
Although professional translators tend to perform ``only the necessary and sufficient corrections"~\cite{bojar-etal-2015}, the validity of test data created by professional translators, including the WMT 2019 test data set, can also be disputable because other native speakers might argue that they can perform better postediting.
For example, some people may consider hyphenated compound ``\textit{Zoom-Werkzeug}'' (`Zoom tool') more natural than closed compound ``\textit{Zoomwerkzeug}'' (Table~\ref{tab:case-2}).

However, considering the big differences in the proportion of \textsc{Negl} (2.35\textit{pp}), which is the set of the cases where an APE system neglects to postedit the given MT, and the F1 score (Table~\ref{tab:frequencies}), it appears that such a risk need not be considered in this analysis.
Moreover, the proposed model has fewer \textsc{Ruin} cases (1.56\%), where it injects errors to the given, already perfect MT, than the baseline model (1.86\%).
Although the proposed model has more \textsc{Degr} cases (7.33\%), where it degrades the given MT, than the baseline (6.65\%), the proposed model's quality degradation $\mu_{\delta \textsc{Bleu}} = -\text{11.72}$ is less severe than that of the baseline ($\mu_{\delta \textsc{Bleu}} = -\text{13.51}$).
Therefore, we conclude that the proposed method results in small but certain improvements.

\section{Conclusion}

To improve the APE quality for high-quality MTs, we propose a linguistically motivated method of regularization that enhances Transformer-based APE models' understanding of the target language: a loss function that encourages APE models to perform symmetric self-attention on a given MT.
Experimental results suggest that the proposed method helps improving the state-of-the-art architecture's APE quality for high-quality MTs; we also present a relative frequency distribution of successes and failures in APE and see increases in the proportion of perfect postediting and the F1 score.
This evaluation method could be useful for assessing the APE quality for high-quality MTs in general.
Actual cases support that the proposed method successfully instills the idea of syntactic symmetry into APE models.
Future research should consider different language pairs and different sets of hyperparameters.

\section{Acknowledgements}

This work was supported by Institute of Information \& Communications Technology Planning \& Evaluation (IITP) grant funded by the Korean government (Ministry of Science and ICT) (No. 2019-0-01906, Artificial Intelligence Graduate School Program (POSTECH)).
We thank Richard Albrecht for assistance in the manual categorization of cases.

\section{Limitations}
First, neither \textit{Feldermodell}~\cite{reis-1980, woellstein-2018, hoehle-2019} nor \textit{Doppelbaum}~\cite{woellstein-2018} has obtained complete concurrence among linguists.
Also, we limit our scope to the English--German language pair and the IT domain using the WMT 2019 training, validation, and test data sets.
A broader scope would not provide confidence in the validity of conducted experiments because there are hardly any standard setups for experimental research~\cite{chatterjee-etal-2018, chatterjee-etal-2019, akhbardeh-etal-2021}.

In addition, the conducted experiment should take into consideration the effect of randomness that is attended in the process of training artificial neural networks; different techniques, different hyperparameters, and multiple runs of optimizers~\cite{clark-etal-2011} may present different results.
However, as previous studies~\cite{chatterjee-etal-2018, chatterjee-etal-2019, chatterjee-etal-2020, akhbardeh-etal-2021}, including the study on the baseline model~\cite{shin-etal-2021}, do not consider the effect of randomness, we also do not investigate the effect of randomness further, considering that training multiple models (Appendix~\ref{sec:exp}) to obtain good estimators (TER and \textsc{Bleu}) will cost a lot.

\bibliography{anthology,custom}

\begin{thebibliography}{24}
\expandafter\ifx\csname natexlab\endcsname\relax\def\natexlab#1{#1}\fi

\bibitem[{Akhbardeh et~al.(2021)Akhbardeh, Arkhangorodsky, Biesialska, Bojar,
  Chatterjee, Chaudhary, Costa-jussa, Espa{\~n}a-Bonet, Fan, Federmann,
  Freitag, Graham, Grundkiewicz, Haddow, Harter, Heafield, Homan, Huck,
  Amponsah-Kaakyire, Kasai, Khashabi, Knight, Kocmi, Koehn, Lourie, Monz,
  Morishita, Nagata, Nagesh, Nakazawa, Negri, Pal, Tapo, Turchi, Vydrin, and
  Zampieri}]{akhbardeh-etal-2021}
Farhad Akhbardeh, Arkady Arkhangorodsky, Magdalena Biesialska, Ond{\v{r}}ej
  Bojar, Rajen Chatterjee, Vishrav Chaudhary, Marta~R. Costa-jussa, Cristina
  Espa{\~n}a-Bonet, Angela Fan, Christian Federmann, Markus Freitag, Yvette
  Graham, Roman Grundkiewicz, Barry Haddow, Leonie Harter, Kenneth Heafield,
  Christopher Homan, Matthias Huck, Kwabena Amponsah-Kaakyire, Jungo Kasai,
  Daniel Khashabi, Kevin Knight, Tom Kocmi, Philipp Koehn, Nicholas Lourie,
  Christof Monz, Makoto Morishita, Masaaki Nagata, Ajay Nagesh, Toshiaki
  Nakazawa, Matteo Negri, Santanu Pal, Allahsera~Auguste Tapo, Marco Turchi,
  Valentin Vydrin, and Marcos Zampieri. 2021.
\newblock \href {https://aclanthology.org/2021.wmt-1.1} {{Findings of the 2021
  Conference on Machine Translation (WMT21)}}.
\newblock In \emph{Proceedings of the Sixth Conference on Machine Translation},
  pages 1--88, Online. Association for Computational Linguistics.

\bibitem[{Bojar et~al.(2015)Bojar, Chatterjee, Federmann, Haddow, Huck, Hokamp,
  Koehn, Logacheva, Monz, Negri, Post, Scarton, Specia, and
  Turchi}]{bojar-etal-2015}
Ond{\v{r}}ej Bojar, Rajen Chatterjee, Christian Federmann, Barry Haddow,
  Matthias Huck, Chris Hokamp, Philipp Koehn, Varvara Logacheva, Christof Monz,
  Matteo Negri, Matt Post, Carolina Scarton, Lucia Specia, and Marco Turchi.
  2015.
\newblock \href {https://doi.org/10.18653/v1/W15-3001} {{Findings of the 2015
  Workshop on Statistical Machine Translation}}.
\newblock In \emph{Proceedings of the Tenth Workshop on Statistical Machine
  Translation}, pages 1--46, Lisbon, Portugal. Association for Computational
  Linguistics.

\bibitem[{Chatterjee et~al.(2019)Chatterjee, Federmann, Negri, and
  Turchi}]{chatterjee-etal-2019}
Rajen Chatterjee, Christian Federmann, Matteo Negri, and Marco Turchi. 2019.
\newblock \href {https://doi.org/10.18653/v1/W19-5402} {{Findings of the WMT
  2019 Shared Task on Automatic Post-Editing}}.
\newblock In \emph{Proceedings of the Fourth Conference on Machine Translation
  (Volume 3: Shared Task Papers, Day 2)}, pages 11--28, Florence, Italy.
  Association for Computational Linguistics.

\bibitem[{Chatterjee et~al.(2020)Chatterjee, Freitag, Negri, and
  Turchi}]{chatterjee-etal-2020}
Rajen Chatterjee, Markus Freitag, Matteo Negri, and Marco Turchi. 2020.
\newblock \href {https://aclanthology.org/2020.wmt-1.75} {{Findings of the WMT
  2020 Shared Task on Automatic Post-Editing}}.
\newblock In \emph{Proceedings of the Fifth Conference on Machine Translation},
  pages 646--659, Online. Association for Computational Linguistics.

\bibitem[{Chatterjee et~al.(2018)Chatterjee, Negri, Rubino, and
  Turchi}]{chatterjee-etal-2018}
Rajen Chatterjee, Matteo Negri, Raphael Rubino, and Marco Turchi. 2018.
\newblock \href {https://doi.org/10.18653/v1/W18-6452} {{Findings of the WMT
  2018 Shared Task on Automatic Post-Editing}}.
\newblock In \emph{Proceedings of the Third Conference on Machine Translation:
  Shared Task Papers}, pages 710--725, Belgium, Brussels. Association for
  Computational Linguistics.

\bibitem[{Clark et~al.(2011)Clark, Dyer, Lavie, and Smith}]{clark-etal-2011}
Jonathan~H. Clark, Chris Dyer, Alon Lavie, and Noah~A. Smith. 2011.
\newblock \href {https://aclanthology.org/P11-2031} {{Better Hypothesis Testing
  for Statistical Machine Translation: Controlling for Optimizer Instability}}.
\newblock In \emph{Proceedings of the 49th Annual Meeting of the Association
  for Computational Linguistics: Human Language Technologies}, pages 176--181,
  Portland, Oregon, USA. Association for Computational Linguistics.

\bibitem[{H{\"o}hle(2019)}]{hoehle-2019}
Tilman~N. H{\"o}hle. 2019.
\newblock \href {https://doi.org/10.5281/zenodo.2588383} {{Topologische
  Felder}}.
\newblock In Stefan M{\"u}ller, Marga Reis, and Frank Richter, editors,
  \emph{Beitr{\"a}ge zur deutschen Grammatik: Gesammelte Schriften von Tilman
  N. H{\"o}hle}, 2 edition, volume~5 of \emph{Classics in Linguistics}, pages
  7--90. Language Science Press, Berlin, Germany.

\bibitem[{Kingma and Ba(2015)}]{kingma-ba-2015}
Diederik~P. Kingma and Jimmy Ba. 2015.
\newblock \href {http://arxiv.org/abs/1412.6980} {{Adam: A Method for
  Stochastic Optimization}}.
\newblock In \emph{3rd International Conference on Learning Representations,
  ICLR 2015, San Diego, CA, USA, May 7-9, 2015, Conference Track Proceedings}.

\bibitem[{Klein et~al.(2017)Klein, Kim, Deng, Senellart, and
  Rush}]{klein-etal-2017}
Guillaume Klein, Yoon Kim, Yuntian Deng, Jean Senellart, and Alexander Rush.
  2017.
\newblock \href {https://aclanthology.org/P17-4012} {{OpenNMT: Open-Source
  Toolkit for Neural Machine Translation}}.
\newblock In \emph{Proceedings of ACL 2017, System Demonstrations}, pages
  67--72, Vancouver, Canada. Association for Computational Linguistics.

\bibitem[{Knight and Chander(1994)}]{knight-chander-1994}
Kevin Knight and Ishwar Chander. 1994.
\newblock \href {https://cdn.aaai.org/AAAI/1994/AAAI94-119.pdf} {{Automated
  Postediting of Documents}}.
\newblock In \emph{Proceedings of the AAAI Conference on Artificial
  Intelligence, 12}, pages 779--784.

\bibitem[{Koehn et~al.(2007)Koehn, Hoang, Birch, Callison-Burch, Federico,
  Bertoldi, Cowan, Shen, Moran, Zens, Dyer, Bojar, Constantin, and
  Herbst}]{koehn-etal-2007}
Philipp Koehn, Hieu Hoang, Alexandra Birch, Chris Callison-Burch, Marcello
  Federico, Nicola Bertoldi, Brooke Cowan, Wade Shen, Christine Moran, Richard
  Zens, Chris Dyer, Ond{\v{r}}ej Bojar, Alexandra Constantin, and Evan Herbst.
  2007.
\newblock \href {https://aclanthology.org/P07-2045} {{Moses: Open Source
  Toolkit for Statistical Machine Translation}}.
\newblock In \emph{Proceedings of the 45th Annual Meeting of the Association
  for Computational Linguistics Companion Volume Proceedings of the Demo and
  Poster Sessions}, pages 177--180, Prague, Czech Republic. Association for
  Computational Linguistics.

\bibitem[{Mitchell(1980)}]{mitchell-1980}
Tom~M. Mitchell. 1980.
\newblock \href {https://www.cs.cmu.edu/~tom/pubs/NeedForBias_1980.pdf} {{The
  Need for Biases in Learning Generalizations}}.
\newblock Technical report, Rutgers University, New Brunswick, NJ.

\bibitem[{Negri et~al.(2018)Negri, Turchi, Chatterjee, and
  Bertoldi}]{negri-etal-2018}
Matteo Negri, Marco Turchi, Rajen Chatterjee, and Nicola Bertoldi. 2018.
\newblock \href {https://aclanthology.org/L18-1004} {{eSCAPE: a Large-scale
  Synthetic Corpus for Automatic Post-Editing}}.
\newblock In \emph{Proceedings of the Eleventh International Conference on
  Language Resources and Evaluation (LREC 2018)}, pages 24--30, Miyazaki,
  Japan. European Language Resources Association (ELRA).

\bibitem[{Papineni et~al.(2002)Papineni, Roukos, Ward, and
  Zhu}]{papineni-etal-2002}
Kishore Papineni, Salim Roukos, Todd Ward, and Wei-Jing Zhu. 2002.
\newblock \href {https://doi.org/10.3115/1073083.1073135} {{BLEU: a Method for
  Automatic Evaluation of Machine Translation}}.
\newblock In \emph{Proceedings of the 40th Annual Meeting on Association for
  Computational Linguistics}, pages 311--318, Philadelphia, Pennsylvania, USA.
  Association for Computational Linguistics.

\bibitem[{Pappas et~al.(2018)Pappas, Miculicich, and
  Henderson}]{pappas-etal-2018}
Nikolaos Pappas, Lesly Miculicich, and James Henderson. 2018.
\newblock \href {https://doi.org/10.18653/v1/W18-6308} {{Beyond Weight Tying:
  Learning Joint Input-Output Embeddings for Neural Machine Translation}}.
\newblock In \emph{Proceedings of the Third Conference on Machine Translation:
  Research Papers}, pages 73--83, Brussels, Belgium. Association for
  Computational Linguistics.

\bibitem[{Reis(1980)}]{reis-1980}
Marga Reis. 1980.
\newblock \href {https://www.persee.fr/doc/drlav_0754-9296_1980_num_22_1_957}
  {{On Justifying Topological Frames : `Positional Field' and the Order of
  Nonverbal Constituents in German ⁰}}.
\newblock \emph{Documentation et Recherche en Linguistique Allemande
  Vincennes}, 22-23:59--85.

\bibitem[{Sennrich et~al.(2016)Sennrich, Haddow, and
  Birch}]{sennrich-etal-2016}
Rico Sennrich, Barry Haddow, and Alexandra Birch. 2016.
\newblock \href {https://doi.org/10.18653/v1/P16-1162} {{Neural Machine
  Translation of Rare Words with Subword Units}}.
\newblock In \emph{Proceedings of the 54th Annual Meeting of the Association
  for Computational Linguistics (Volume 1: Long Papers)}, pages 1715--1725,
  Berlin, Germany. Association for Computational Linguistics.

\bibitem[{Shin et~al.(2021)Shin, Lee, Go, Jung, Kim, and Lee}]{shin-etal-2021}
Jaehun Shin, Wonkee Lee, Byung-Hyun Go, Baikjin Jung, Youngkil Kim, and
  Jong-Hyeok Lee. 2021.
\newblock \href {https://doi.org/10.1145/3465383} {{Exploration of Effective
  Attention Strategies for Neural Automatic Post-Editing with Transformer}}.
\newblock \emph{ACM Transactions on Asian and Low-Resource Language Information
  Processing}, 20(6).

\bibitem[{Snover et~al.(2006)Snover, Dorr, Schwartz, Micciulla, and
  Makhoul}]{snover-etal-2006}
Matthew Snover, Bonnie Dorr, Rich Schwartz, Linnea Micciulla, and John Makhoul.
  2006.
\newblock \href {https://aclanthology.org/2006.amta-papers.25} {{A Study of
  Translation Edit Rate with Targeted Human Annotation}}.
\newblock In \emph{Proceedings of the 7th Conference of the Association for
  Machine Translation in the Americas: Technical Papers}, pages 223--231,
  Cambridge, Massachusetts, USA. Association for Machine Translation in the
  Americas.

\bibitem[{Srivastava et~al.(2014)Srivastava, Hinton, Krizhevsky, Sutskever, and
  Salakhutdinov}]{srivastava-etal-2014}
Nitish Srivastava, Geoffrey Hinton, Alex Krizhevsky, Ilya Sutskever, and Ruslan
  Salakhutdinov. 2014.
\newblock \href {http://jmlr.org/papers/v15/srivastava14a.html} {{Dropout: A
  Simple Way to Prevent Neural Networks from Overfitting}}.
\newblock \emph{Journal of Machine Learning Research}, 15(56):1929--1958.

\bibitem[{Sutskever et~al.(2014)Sutskever, Vinyals, and
  Le}]{sutskever-etal-2014}
Ilya Sutskever, Oriol Vinyals, and Quoc~V Le. 2014.
\newblock \href
  {https://proceedings.neurips.cc/paper_files/paper/2014/file/a14ac55a4f27472c5d894ec1c3c743d2-Paper.pdf}
  {{Sequence to Sequence Learning with Neural Networks}}.
\newblock In \emph{Advances in Neural Information Processing Systems},
  volume~27. Curran Associates, Inc.

\bibitem[{Vaswani et~al.(2017)Vaswani, Shazeer, Parmar, Uszkoreit, Jones,
  Gomez, Kaiser, and Polosukhin}]{vaswani-etal-2017}
Ashish Vaswani, Noam Shazeer, Niki Parmar, Jakob Uszkoreit, Llion Jones,
  Aidan~N. Gomez, {\L}ukasz Kaiser, and Illia Polosukhin. 2017.
\newblock \href
  {https://proceedings.neurips.cc/paper/2017/file/3f5ee243547dee91fbd053c1c4a845aa-Paper.pdf}
  {{Attention is All you Need}}.
\newblock In \emph{Advances in Neural Information Processing Systems},
  volume~30. Curran Associates, Inc.

\bibitem[{W{\"o}llstein(2018)}]{woellstein-2018}
Angelika W{\"o}llstein. 2018.
\newblock \href
  {https://ids-pub.bsz-bw.de/frontdoor/deliver/index/docId/7998/file/Woellstein_Topologisches_Satzmodell_2018.pdf}
  {{Topologisches Satzmodell}}.
\newblock In J{\"o}rg Hagemann and Sven Staffeldt, editors,
  \emph{Syntaxtheorien: Analysen im Vergleich}, volume~28 of \emph{Stauffenburg
  Einf{\"u}hrungen}, pages 145--166. Stauffenburg, Tübingen, Germany.

\bibitem[{Yang et~al.(2018)Yang, Huang, and Ma}]{yang-etal-2018}
Yilin Yang, Liang Huang, and Mingbo Ma. 2018.
\newblock \href {https://doi.org/10.18653/v1/D18-1342} {{Breaking the Beam
  Search Curse: A Study of (Re-)Scoring Methods and Stopping Criteria for
  Neural Machine Translation}}.
\newblock In \emph{Proceedings of the 2018 Conference on Empirical Methods in
  Natural Language Processing}, pages 3054--3059, Brussels, Belgium.
  Association for Computational Linguistics.

\end{thebibliography}
\bibliographystyle{acl_natbib}

\appendix

\section{Experimental Details}
\label{sec:exp}

We use the following hyperparameters: the number of layers $N = \text{6}$, the number of heads $H = \text{8}$, the dimension of key vectors $d_{k} = \text{64}$, the dimension of value vectors $d_{v} = \text{64}$, the vector dimension for multi-head attention layers $d_{\text{model}} = \text{512}$, the vector dimension for the inner layer of position-wise feed-forward networks $d_{\text{ff}} = \text{2,048}$, the dropout~\cite{srivastava-etal-2014} probability $P_{\text{drop}} = 0.1$, the label smoothing value $\epsilon_{\text{LS}} = 0.1$, minibatches of 25,000 tokens, a learning rate of $2.0$, warmup for 18,000 training steps, and a shared vocabulary consisting of 32,000 subword units~\cite{sennrich-etal-2016}\footnote{
    We used \href{https://github.com/google/sentencepiece}{SentencePiece} (Apache License 2.0)
}.
We also use weight tying~\cite{pappas-etal-2018} and the Adam optimizer~\cite{kingma-ba-2015} with $\beta_{1}=0.9$, $\beta_{2}=0.998$, and $\epsilon=10^{-8}$.
Decoding options are beam search with a beam size $b =$ 5, a length penalty multiplied by a strength coefficient $\alpha = 0.6$, and beam search stopping~\cite{yang-etal-2018} with the length ratio $lr = 1.3$.

We use \href{https://github.com/OpenNMT/OpenNMT-py}{OpenNMT-py} 3.0~\cite{klein-etal-2017}\footnote{The MIT License.} with the random seed $1128$.
We first train the models for 100,000 steps, about 36 hours on one NVIDIA GeForce RTX\textsuperscript{\texttrademark} 3090, and then tune them around 1,000 steps.

\end{document}